\documentclass[a4paper, 11pt, draftclsnofoot, onecolumn]{IEEEtran}

\usepackage{amsmath, amssymb}

\usepackage{graphicx}               

\newcommand{\eqdef}{\stackrel{\rm def}{=}}

\newcommand{\bg}{\mathbf{g}}

\newcommand{\bn}{\mathbf{n}}

\newcommand{\bs}{\mathbf{s}}

\newcommand{\bw}{\mathbf{w}}
\newcommand{\bx}{\mathbf{x}}

\newcommand{\bC}{\mathbf{C}}

\newcommand{\bH}{\mathbf{H}}
\newcommand{\bI}{\mathbf{I}}

\newcommand{\bR}{\mathbf{R}}

\newcommand{\bW}{\mathbf{W}}

\renewcommand{\t}{{^{\mathrm{T}}}} 
\newcommand{\tc}{^{\mathrm{H}}} 
\newcommand{\E}{\mathrm{E}}

\newcommand{\cK}{\mathcal{K}}
\newcommand{\cM}{\mathcal{M}}

\newcommand{\enleve}[1]{\relax}

\newcommand{\re}{\text{I$\!$Re}}    
\newcommand{\opt}{_{\mathrm{opt}}}

\title{Robust Independent Component Analysis by Iterative Maximization
  of the Kurtosis Contrast with Algebraic Optimal Step Size}

\author{Vicente Zarzoso, {\it Member, IEEE}, and Pierre Comon, {\it
    Fellow, IEEE}\thanks{Manuscript submitted
    March~11, 2009; revised October~13, 2009; accepted October 24, 2009.}\thanks{The authors are with
    the I3S Laboratory, University of Nice - Sophia Antipolis,
    CNRS, Les Algorithmes, Euclide-B, BP 121, 2000 route des
    Lucioles, 06903 Sophia Antipolis Cedex, France. {\tt {\{zarzoso, pcomon\}@i3s.unice.fr}}.}}

\begin{document}

\maketitle
\begin{abstract}
Independent component analysis (ICA) aims at decomposing an observed
random vector into statistically independent
variables. Deflation-based implementations, such as the popular
one-unit FastICA algorithm and its variants, extract the independent components one after
another. A novel method for deflationary ICA, referred to as
RobustICA, is put forward in this paper. This simple technique consists of
performing exact line search optimization of the kurtosis
contrast function. The step size leading to the global maximum of the contrast
along the search direction is found among the roots of a fourth-degree
polynomial. This polynomial rooting can be performed algebraically, and thus at low cost, at each iteration. Among other practical benefits, RobustICA can avoid
prewhitening and deals with real- and complex-valued mixtures of possibly
non-circular sources alike. The absence of prewhitening improves
asymptotic performance. The algorithm is robust to local
extrema and shows a very high convergence speed in
terms of the computational cost required to reach a given source extraction
quality, particularly for short data
records. These features are demonstrated by a comparative numerical
analysis on synthetic data. RobustICA's capabilities in
processing real-world data involving non-circular complex strongly
super-Gaussian sources are
illustrated by the biomedical problem of atrial activity (AA) extraction
in atrial fibrillation (AF) electrocardiograms (ECGs), where it
outperforms an alternative ICA-based technique.
\end{abstract}

\begin{keywords}
Atrial fibrillation (AF), blind source separation (BSS), independent
component analysis (ICA), iterative
optimization, kurtosis, optimal step size, performance analysis.
\end{keywords}


\section{Introduction}\label{sec:intro}


\subsection{Blind Source Separation and Independent Component Analysis}

Introduced over two
decades ago \cite{HER85}, the problem of blind source separation (BSS) consists of recovering a set of unobservable
source signals from observed mixtures of the sources.  
Independent component analysis (ICA) aims at decomposing an observed
random vector into statistically independent variables
\cite{COM94}.  Among its numerous applications, ICA is the most natural
tool for BSS in instantaneous linear
mixtures when the source signals are assumed to be independent. As
opposed to classical decomposition techniques such as principal component
analysis (PCA), ICA can deal with a general mixing structure, even if not made up of
orthogonal columns.
The plausibility of the statistical
independence assumption in a wide variety of fields,
including telecommunications, finance and
biomedical engineering, helps explain the
arousing interest in this research area witnessed over the last two
decades. 

Mathematically, the observed random vector $\bx \in \mathbb{C}^L$ is assumed to be
generated according to the instantaneous linear mixing model:
\begin{equation} \label{eq:model}
\bx=\bH\bs + \bn
\end{equation}
where the source vector $\bs = [s_1, s_2, \dots, s_K]\t \in
\mathbb{C}^K$ is made of $K \le L$
unknown mutually independent components. The elements of mixing matrix $\bH \in
\mathbb{C}^{L\times K}$ are also unknown, and so are the noise vector
$\bn$ and its probability distribution; the noise is only assumed to
be independent of the sources. Our focus is on batch or block implementations, which, contrary to common belief, are
not necessarily more costly than adaptive (recursive, on-line,
sample-by-sample, or neural) algorithms, and are able to use more effectively the
information contained in the observed signal block \cite{AMA98nc}. Given a sensor-output signal block composed
of $T$ samples, ICA aims at estimating the corresponding $T$-sample realization of
the source vector.


\subsection{Kurtosis as a Contrast Function}

Since Comon's seminal work \cite{COM94}, many contrast functions for ICA have been
proposed in the literature, mainly based on information theoretical
principles such as maximum likelihood, mutual information, marginal
entropy and negentropy, as well as related
non-Gaussianity measures \cite{PHA97, CAR98, CAR99}. 
Among them, the kurtosis
(normalized fourth-order marginal cumulant) is arguably the most common
statistics used in ICA, even if skewness has also been proposed
\cite{COM94ifac}. The use of kurtosis dates back to the work of Wiggins \cite{WIG78}, Donoho \cite{DON80} and Shalvi-Weinstein
\cite{SHA90} on blind deconvolution of seismic signals and blind
equalization of single-input single-output (SISO) digital
communication channels, two problems that can be related to BSS/ICA.  
 One of the main benefits of kurtosis lies in the absence of spurious
local extrema for infinite sample size when the noiseless observation model is fulfilled. This attractive feature leads to {\em globally
convergent} source extraction algorithms, from which full source
separation can be performed by using some form of deflation procedure
\cite{DEL95, HYV97, DIN00, PAP00}, even in the convolutive MIMO case \cite{TUG97}.
Although the adequacy of kurtosis as a contrast may be objected on the
basis of statistical efficiency and robustness against outliers
\cite{HYV97nnsp}, its widespread use is justified by
mathematical tractability, computational convenience and robustness to finite sample
effects. Theoretical evidence for its finite-sample robustness have been
gathered by previous works. In \cite{BER05}, the sample kurtosis yields an estimate with less
variance than the fourth-order moment and the fourth-order cumulant
for all distributions tested, including sub-Gaussian and
super-Gaussian densities. As an extension of these results, using
the full expression of the fourth-order cumulant instead of the
simplified form employed, e.g., in the FastICA algorithm \cite{HYV97,
  HYV99} is shown to improve extraction performance
\cite{BER07}.  The computational convenience and finite sample
robustness of kurtosis can be further improved by the optimal
step-size iterative search proposed in the present paper. In the
presence of outliers, the performance
of the conventional kurtosis estimate based on sample moments can be enhanced by means
of more robust alternative estimates available in the literature (see,
e.g., \cite[Ch.~5]{NAN99}).


\subsection{The FastICA Algorithm}

The FastICA algorithm \cite{HYV97, HYV97nnsp, HYV99, HYV01} is 
perhaps the most popular method for ICA, due to its simplicity,
convergence speed and satisfactory results in numerous applications.
Indeed, the one-unit algorithm with cubic non-linearity, related to
the optimization of the
kurtosis contrast under prewhitening, offers cubic
global convergence if the ICA model is fulfilled and the sample size tends to infinity \cite{HYV97, DOU03}. In addition, the algorithm is asymptotically efficient if the
non-linearity is matched to the source probability density function \cite{TIC06}. The cubic
non-linearity associated with kurtosis is particularly well adapted to
sub-Gaussian distributions \cite{HYV97nnsp, TIC06}.
 Some of these
desirable properties are also shared by the symmetric
version of the algorithm \cite{OJA06}.
Originally put
forward in deflation mode, FastICA appeared after other kurtosis-based
ICA methods such as CoM2
\cite{COM94}, JADE \cite{CAR93}, CoM1
\cite{COM97}, or the deflation methods by Tugnait
\cite{TUG97} or Delfosse-Loubaton \cite{DEL95}.
A first comparison with earlier methods can be found in \cite{GIA99}.
In the comparative study of \cite{CHE04ijcnn}, FastICA is shown to
fail for weak or highly spatially correlated sources. Its convergence slows down or even fails in the presence of saddle
points, particularly for short block sizes \cite{TIC06}. To surmount
this difficulty, a simple
saddle-point check method is proposed in that reference. Such a method is based
on estimated
component pairs and, as a result, is not applicable if only one independent component is required. 
Further improvements of the symmetric implementation of the algorithm
are developed in \cite{KOL06}. All these results rely heavily on the
assumption that the observed signals have been perfectly whitened
or sphered before further higher-order processing. 
As pointed out in \cite{CAR94}, the use of prewhitening imposes a
 bound on separation performance and introduces an estimation bias due to
residual source correlations for short data sizes.

\subsection{The Complex-Valued Scenario}

The FastICA algorithm was originally developed for real-valued
signals only. A first extension to complex-valued sources is proposed in \cite{BIN00}, and later
shown to keep the cubic global
convergence property of its real counterpart \cite{RIS02}. Such an extension, however, is only valid for second-order
circular sources, a limitation that has motivated more recent efforts to extend the usefulness of the algorithm to non-circular
sources \cite{DOU07, NOV08, NOV08tnn, LI08}. Reference \cite{LI08} derives gradient, fixed-point and Newton-like
algorithms based on the general definition of the fourth-order marginal cumulant valid for
non-circular sources. In \cite{NOV08} the
whitened observation pseudo-covariance matrix is incorporated into FastICA's
update rule to guarantee local stability
at the separating solutions even in the presence of non-circular
sources. For the kurtosis-based non-linearity, the resulting algorithm
bears close resemblance to that derived in \cite{DOU07} through an
ingenious approach sparing
differentiation. Similar algorithms are proposed in \cite{NOV08tnn}
through a negentropy-based family of cost functions preserving phase
information and thus adapted to non-circular sources. Such functions
must be chosen in accordance with the source distributions to assure
stability. Again, all the above methods rely on prewhitening. 
Interestingly, early methods for BSS in the complex case did not
require prewhitening and were also applicable to
non-circular sources~\cite{MOR94, MOR94eusipco}.

\subsection{Summary and Contributions of the Paper}

This contribution presents a novel method for
deflationary ICA named RobustICA \cite{eusipco06, ica07, embc08}. The
method is based on a general contrast function, the kurtosis, which is optimized by a
computationally efficient technique based on an optimal step size
(adaption coefficient). Any independent component with
non-zero kurtosis can be extracted in this manner. No simplifying
assumptions concerning specific type
of sources (real or complex, circular or non-circular, sub-Gaussian or
super-Gaussian) are involved in the
derivation of the algorithm. The
methodology behind RobustICA is exact line search, well known in the field
of numerical optimization (see, e.g.,
\cite{PRE92}). However, classical line search techniques can only perform
iterative local optimization along the search direction. By contrast, the optimal
step-size technique used in RobustICA computes {\em algebraically}
(i.e., without iterations) the
step size {\em globally} optimizing the kurtosis in
the search direction at each extracting vector update. When compared to other
kurtosis-based algorithms such as the original FastICA and its variants, the method
presents a number of advantages with significant practical impact:
\begin{itemize}

\item As opposed to~\cite{HYV99, BIN00, RIS02} and related works, the generality of the kurtosis contrast guarantees that real-
  and complex-valued signals can be treated by exactly the same
  algorithm without any modification. Both type of source signals can be present simultaneously
  in a given mixture, and complex sources need not be circular. The mixing matrix coefficients may be real or complex, regardless of the source type.

\item Contrary to most ICA methods, prewhitening is not required, so that the performance
  limitations it imposes \cite{CAR94} can be avoided. Sequential extraction
  (deflation) can be carried out, e.g., via linear
  regression. This feature may
  prove especially beneficial in ill-conditioned scenarios, the
  convolutive case and underdetermined mixtures.\footnote{Other BSS
 methods avoiding prewhitening or dealing with
  non-circular complex sources have been proposed elsewhere in the literature.}

\item The algorithm can target sub-Gaussian or
  super-Gaussian sources in the order specified by the user. This
feature enables the extraction of sources of interest when their
Gaussianity character is known in advance, thus sparing a full
separation of the observed mixture as well as the consequent increased
  complexity and estimation error.

\item The optimal step-size technique provides some robustness to the
  presence of saddle points and spurious local extrema in the contrast
  function.

\item The method shows a very high convergence speed measured in terms
  of source extraction quality versus number of operations. In the
  real-valued two-signal case, the algorithm converges in a single
  iteration, even without prewhitening.

\end{itemize}
RobustICA's cost-efficiency and robustness are particularly remarkable
for short sample length in the absence of prewhitening.
In addition to presenting the method and assessing its comparative performance
 on synthetic data, the practical usefulness of RobustICA is
 illustrated in a real-world problem: the extraction of the
 atrial activity signal from surface electrocardiogram (ECG) recordings of
 atrial fibrillation. This biomedical application demonstrates that the kurtosis
 contrast can also be used with success in the extraction of strongly
 super-Gaussian sources, which, in addition, present non-circular
 complex distributions in this particular context.

\subsection{Related Work on Optimal Step-Size Iterative Methods}

The convergence properties of
iterative techniques are to a large extent
determined by the step size, learning rate or adaption coefficient
employed in their update equations. It is well known that the step-size choice sets a
difficult balance between convergence speed and final accuracy (misadjustment). 
This trade-off has spurred
the development of iterative techniques based on some form of
step-size optimization. 
To our knowledge, research into adaptive step-size optimization can be
traced back to the work of Kuzminskiy on the least mean squares (LMS) algorithm in nonstationary
environments, where recursive expressions for the step
size are derived  \cite{KUZ82,
  KUZ97dsp}. More recent works on the
LMS algorithm such as \cite{HUA05, GAU06} seem
closer to our approach, except that they aim at channel identification
 and the optimal step-size is computed
using a quadratic cost function different from that minimized via the stochastic
LMS. Our rationale is essentially different, as we aim at direct
source estimation and globally optimize
a non-quadratic contrast by iterating on the same signal block under the assumption of stationarity over
the observation window (block or batch processing).

Amari~\cite{AMA98nc, AMA98} puts forward adaptive rules for learning
the step size in neural algorithms for BSS/ICA, more pertinent in the context
of the present work. The idea is to make
the step size depend on the gradient norm, in order to obtain a fast evolution at the
beginning of the iterations and then a decreasing misadjustment as a
stationary point is reached. These step-size learning rules, in turn,
include other learning coefficients which must be set
appropriately. Although the resulting algorithms are said to be robust
to the choice of these coefficients, their optimal selection remains
application dependent. Other guidelines for choosing the step size in
natural gradient algorithms are given in \cite{CRU04}, but are merely
based on local stability conditions. In a non-linear mixing setup,
Khor and co-workers put forward a fuzzy logic approach to control the learning rate of a
separation algorithm based on the natural gradient \cite{KHO05}.

In the context of batch algorithms, Regalia~\cite{REG02} finds bounds for the step size
guaranteeing monotonic convergence of the normalized fourth-order
moment of the extractor output. Such a functional is only a contrast for real-valued sources under
prewhitening, a similar limitation shared by the more general class of
functions considered in \cite{REG03}. Determining these
step-size bounds is a computational intensive task, as it involves the
eigenspectrum of a Hessian matrix on a convex subset
containing the unit sphere in the $K$-dimensional space. While still ensuring monotonic convergence, 
the optimal step-size approach that we develop herein is valid for real- and
complex-valued sources, does not require prewhitening and is
computationally very simple.
This type of technique has already been successfully applied by the
authors to other higher order contrasts such as the constant
modulus or the constant power criteria in the problems of blind and semi-blind
equalization of digital communication channels \cite{eusipco05, icassp05, tsp05, tcom08}.

\subsection{Organization of the Paper}

The paper begins by critically reviewing the deflationary kurtosis-based FastICA
algorithm and its variants in
Sec.~\ref{sec:fastica}. Then, Sec.~\ref{sec:robustica} presents the
RobustICA technique. Its experimental comparative assessment is
carried out in Sec.~\ref{sec:exp}. In particular, we aim at evaluating objectively the
algorithms' speed and efficiency by taking into account the cost per
iteration in number of operations. A biomedical application, the
extraction of atrial activity from ECG recordings of atrial
fibrillation, illustrates the method's
ability to deal with non-circular complex-valued super-Gaussian sources, as reported in
Sec.~\ref{sec:aaextr}. The
concluding remarks of Sec.~\ref{sec:con} bring the paper to an end.


\section{FastICA Revisited} \label{sec:fastica}


\subsection{Kurtosis-Based Optimality Criteria} \label{sec:criteria}

In the deflation approach to ICA, an extracting vector $\bw$ is sought so
that the estimate
\begin{equation} \label{eq:output}
y\eqdef \bw\tc\bx
\end{equation}
where $(\cdot)\tc$ denotes the conjugate-transpose operator, maximizes some optimality criterion or contrast function, and is hence
expected to be a component independent from the others. A widely
used contrast is the kurtosis,
which is defined as the normalized fourth-order marginal cumulant:
\begin{equation} \label{eq:kurt}
\cK(\bw) = \frac{\E\{|y|^4\}-2\E^2\{|y|^2\}-|\E\{y^2\}|^2}{\E^2\{|y|^2\}}
\end{equation}
where $\E\{\cdot\}$ denotes the mathematical expectation.
This criterion is easily seen to be insensitive to scale, i.e.,
$\cK(\lambda\bw)=\cK(\bw)$, $\forall\lambda\neq0$. Since this scale
indeterminacy is typically unimportant, we can impose, without loss of
generality, the normalization $\|\bw\|=1$ for
numerical convenience. The {\it kurtosis maximization (KM)} criterion based on contrast
(\ref{eq:kurt}) is quite general in that it does not require the
observations to be prewhitened and can be applied to real- or complex-valued
sources without any modification.

The KM criterion started to receive attention with the pioneering work
of Wiggins \cite{WIG78}, Donoho \cite{DON80} and
Shalvi-Weinstein \cite{SHA90} on blind deconvolution, and was later
employed for source separation \cite{DEL95}, even in the convolutive mixture scenario \cite{TUG97}.
In the real-valued case, it was proved in
\cite{DEL95} that the
maximization of criterion $|\cK(\bw)|$ is a valid contrast for the extraction of any source with non-zero kurtosis from
model~(\ref{eq:model}) after prewhitening. To avoid extracting the same
source twice, the remaining unitary mixing matrix is suitably parameterized as a
function of angular parameters, and function~(\ref{eq:kurt}) iteratively maximized
with respect to these angles. In the
convolutive mixture scenario of \cite{TUG97}, the contrast is
maximized without parameterization. Regression
is used as an alternative method to avoid extracting the
same source more than once.
 
To simplify the source extraction, the kurtosis-based FastICA algorithm \cite{HYV97},
\cite{HYV99}, \cite{HYV01} first applies a prewhitening operation, as
in \cite{DEL95},
resulting in transformed observations with an identity covariance
matrix, $\bR_x\eqdef\E\{\bx\bx\tc\} = \bI$. In the real-valued case, contrast (\ref{eq:kurt}) then
becomes equivalent to the fourth-order moment criterion:
\begin{equation}\label{eq:mom}
\cM(\bw) = \E\{|y|^4\}
\end{equation}
which must be optimized under a constraint, e.g., $\|\bw\|=1$, to avoid
arbitrarily large values of $y$. Under the same constraint, criteria (\ref{eq:kurt}) and
(\ref{eq:mom}) are also equivalent if the sources are complex-valued but second-order circular, i.e., the
non-circular second-moment (or pseudo-covariance) matrix $\bC_s\eqdef\E\{\bs\bs\t\}$ is
null, where $(\cdot)\t$ is the transpose operator without conjugation. Consequently, contrast (\ref{eq:mom}) is less general than
criterion (\ref{eq:kurt}) in
that it requires the observations to be prewhitened and the sources
to be real-valued, or complex-valued but circular.


\subsection{Contrast Optimization} \label{sec:opt}

Under the constraint $\|\bw\|=1$, the stationary points of $\cM(\bw)$
are obtained as a collinearity condition on $\E\{yy^{*2}\bx \}$, where
$(\cdot)^*$ denotes complex conjugation: 
\begin{equation} \label{eq:fixedpoint}
\E\bigl\{|\bw\tc\bx|^2\bx\bx\tc\bigr\}\bw=\lambda\bw
\end{equation}
in which $\lambda$ is a Lagrangian multiplier. 
As opposed to the claims of \cite{HYV97},
eqn.~(\ref{eq:fixedpoint}) is a fixed-point equation only if $\lambda$ is known, which is not the
case here; $\lambda$ must be determined so as to satisfy the
constraint, and thus it depends on $\bw\opt$, the optimal value of
$\bw$: $\lambda = \cM(|\bw\opt\tc\bx|^4\}$.

For the sake of simplicity, $\lambda$ is arbitrarily set to a deterministic
fixed value \cite{HYV97}, \cite{HYV01}, so that FastICA
becomes an approximate standard Newton algorithm, as eventually pointed out in \cite{HYV99}. 
In the real-valued case, the Hessian matrix of $\cM(\bw)$ is approximated as
\begin{equation} 
\E\{(\bw\t\bx\bx\t\bw)\bx\bx\t\} \approx
\E\{\bw\t\bx\bx\t\bw\}\E\{\bx\bx\t\} = \bw\t\bw = \bI
\end{equation}
As a result, the kurtosis-based FastICA iteration reduces to
\begin{equation} \label{eq:fastica}
\bw^+ = \bw - \frac{1}{3}\,\E\{\bx(\bw\t\bx)^3\}.
\end{equation}
%
Since $\nabla\cM(\bw) = 4\E\{\bx(\bw\t\bx)^3\}$, eqn.~(\ref{eq:fastica}) is
essentially a 
gradient-descent update rule of the form
\begin{equation*}
\bw^+ = \bw - \mu\nabla\cM(\bw)
\end{equation*}
with a fixed value for the step
size, $\mu = 1/12$. It follows that the kurtosis-based FastICA is a
particular instance, using prewhitening and assuming sub-Gaussian sources, of the family of gradient-based algorithms proposed in
\cite{TUG97}. Though fixed to a constant value, FastICA's step-size
choice is judicious in that it leads to cubic convergence of
the algorithm for infinite sample size \cite{HYV99}. For short sample
sizes, however, convergence may slow down and even get trapped in saddle areas
and local extrema, as has been noticed in \cite{TIC06} and will be
further illustrated in Sec.~\ref{sec:exp}.

To prevent locking onto a previously extracted source, the so-called
deflationary orthogonalization can be performed after
each FastICA update iteration. The extracting vector is constrained to
lie within
the orthogonal subspace of the extracting vectors, stored in matrix
$\bW_k = [\bw_1, \bw_2, \dots, \bw_{k-1}]$, found for the previous
$(k-1)$ sources:
\begin{equation} \label{eq:deflortho}
\bw^+ \leftarrow \bw^+ - \bW_k\bW_k\tc\bw^+.
\end{equation}
This procedure is tantamount to the Gram-Schmidt orthogonalization of
$\bw^+$ with respect to the columns of $\bW_k$.
The iteration concludes with a normalization step to guarantee the
constraint \mbox{$\|\bw^+\| = 1$}:
\begin{equation} \label{eq:norm}
\bw^+ \leftarrow \bw^+/\|\bw^+\|.
\end{equation}
The algorithm can be stopped when
\begin{equation} \label{eq:stop}
\bigl|1 - |\bw\tc\bw^+|\bigr| < \epsilon
\end{equation}
for a statistically significant small constant $\epsilon$, e.g., $\epsilon =
\eta/T$ with $\eta < 1$. The use of the transpose-conjugate operator
in eqns.~(\ref{eq:deflortho}) and~(\ref{eq:stop}) makes them also
valid in the complex case.


\subsection{The Complex Case} \label{sec:complex}

In the
extension of the kurtosis-based FastICA algorithm to complex-valued scenarios \cite{BIN00,
  RIS02}, the update rule can be expressed as
\begin{equation} \label{eq:fasticacmpx}
\bw^+ = \bw - \frac{1}{2}\,\E\{\bx y^*|y|^2\}
\end{equation}
with $y$ given in~(\ref{eq:output}). Let us define the gradient operator as $\nabla_{\bw} = \nabla_{\bw_r} + j
\nabla_{\bw_i}$, where $\bw_r$ and $\bw_i$ represent the real
and imaginary parts, respectively, of vector $\bw$; this a scaled form
of Brandwood's conjugate gradient \cite{BRA83}. Then, 
eqn.~(\ref{eq:fasticacmpx}) is easily shown to be a gradient-descent algorithm on
contrast~(\ref{eq:mom}) with fixed
step size $\mu = 1/8$. The algorithm is
only valid for second-order circular sources, satisfying $\bC_s =
\mathbf{0}$. Recent works aiming to avoid this limitation are all
based on the prewhitening assumption. Starting from the
non-normalized fourth-order cumulant contrast, the KM fixed-point (KM-F) algorithm of
\cite{LI08} assigns the current gradient to the extracting vector
\begin{equation} \label{eq:kmf}
\bw^+ = \E\{|y|^2y^*\bx\} - 2\E\{|y|^2\}\E\{y^*\bx\} - \E\{{y^*}^2\}\E\{y\bx\}
\end{equation}
before the orthogonalization and normalization steps described by
eqns.~(\ref{eq:deflortho}) and~(\ref{eq:norm}). A modification of
\cite{BIN00} is proposed in \cite{NOV08} leading to the so-called non-circular
FastICA (nc-FastICA) algorithm. For contrast~(\ref{eq:mom}), the modified
update rule reads: 
\begin{equation} \label{eq:ncfastica}
\bw^+ = \bw - \frac{1}{2}\,\E\{|y|^2y^*\bx\} + \frac{1}{2}\E\{\bx\bx\t\}\E\{{y^*}^2\}\bw^*.
\end{equation}
By taking into account the
whitened observation pseudo-covariance matrix in the last term, the nc-FastICA
algorithm becomes locally stable at the separation solutions even in
the presence of non-circular sources. 
The complex fixed-point
algorithm (CFPA) of \cite{DOU07} turns out to rely on a very similar
update rule, obtained through an
alternative approach not based on differentiation.


\section{RobustICA} \label{sec:robustica}

\subsection{Exact Line Search on the Kurtosis Contrast} \label{sec:osskurt}

Without simplifying assumptions, a simple quite natural alternative to FastICA consists of performing
exact line search of the absolute kurtosis contrast (\ref{eq:kurt}):  
\begin{equation}\label{eq:exactlinesearch}
\mu\opt = \arg \underset{\mu}{\max}\: |\cK(\bw + \mu\bg)|.
\end{equation}
The search direction $\bg$ is typically (but not necessarily) the
gradient, $\bg = \nabla_\bw\cK(\bw)$, which is given by
(cf.~\cite{DIN00, TUG97}): 
\begin{equation*} 
\nabla_\bw\cK(\bw) = \frac{4}{\E^2\{|y|^2\}}\left\{\E\{|y|^2y^*\bx\} -
  \E\{y\bx\}\E\{{y^*}^2\} - \frac{\bigl(\E\{|y|^4\} -
  |\E\{y^2\}|^2\bigr)\E\{y^*\bx\}}{\E\{|y|^2\}}\right\}.
\end{equation*}
Exact line search is in general computationally intensive and presents
 other limitations \cite{PRE92}, which explains
 why, despite being a well-known optimization method, it is very
 rarely used in practice. Indeed,
 the one-dimensional optimization in eqn.~(\ref{eq:exactlinesearch}) must typically be performed by
 means of numerical algorithms that are not guaranteed to find the
 global optimum along the search direction. However, for criteria that can be expressed
 as polynomials or rational functions of $\mu$, such as the kurtosis, the constant
 modulus \cite{GOD80}, \cite{tcom08} and the constant power
 \cite{GRE98}, \cite{tsp05} contrasts, the {\em globally} optimal step
size $\mu\opt$ can easily be determined {\em algebraically} by finding the roots of a
 low-degree polynomial. The RobustICA algorithm is derived from the
 application of this idea to the kurtosis contrast, as detailed next. A freely available
 Matlab implementation can be found in~\cite{robustica}.

At each iteration, RobustICA performs an optimal step-size (OS) based optimization comprising the following steps:
\begin{itemize}

\item[S1)] Compute the OS polynomial coefficients. 
 
For the kurtosis contrast,
 the OS polynomial is given by:
\begin{equation} \label{eq:pol}
p(\mu) = \sum_{k=0}^4 a_k \mu^k.
\end{equation}
The coefficients $\{a_k\}_{k=0}^4$ can easily can be obtained at each iteration from the observed
 signal block and the current values of $\bw$ and $\bg$. Their
 expressions are found in the Appendix. Numerical conditioning in the determination of $\mu\opt$ can be
 improved by normalizing the gradient vector beforehand.

\item[S2)] Extract OS polynomial roots $\{\mu_k\}_{k=1}^4$.

The roots of the 4th-degree polynomial (quartic) can be found at
practically no cost using standard algebraic procedures such as Ferrari's formula, known since
the 16th century \cite{PRE92}. Indeed, the complexity of
this step is negligible compared with the calculation of the statistics
required in the previous step. Details about computational cost are given in Sec.~\ref{sec:comp}.

\item[S3)] Select the root leading to the absolute maximum of the
  contrast along the search direction:
\begin{equation} \label{eq:optstep}
\mu\opt = \arg \underset{k}{\max}\: |\cK(\bw
+ \mu_k\bg)|.
\end{equation}
This can be done at a negligible cost from the coefficients computed in step~S1, as
detailed in the Appendix.

\item[S4)] Update $\bw^+ = \bw + \mu\opt\bg$.

\item[S5)] Normalize as in eqn.~(\ref{eq:norm}).

\end{itemize}

As in \cite{TUG97}, the extracting vector normalization in step~S5 is
performed to fix the ambiguity introduced by the
the scale invariance of contrast~(\ref{eq:kurt}), and does not stem from prewhitening. 
The same stopping criterion as in FastICA [cf.~eqn.~(\ref{eq:stop})]
can also be employed to check the convergence of the above algorithm.
The generality of contrast~(\ref{eq:kurt}) guarantees
that RobustICA is able to separate real and complex (possibly non-circular) sources
 without any modification. 
These features will be illustrated in the experiments of Secs.~\ref{sec:exp}--\ref{sec:aaextr}.


\subsection{Extraction of Sources with Known Kurtosis Sign} \label{sec:kurtsign}

The method described above aims at maximizing the absolute kurtosis, and is
thus able to extract sources with positive or negative kurtosis. In many applications, some
information may be known in advance about the source(s) of
interest. For example, the atrial activity time-domain signal in atrial
fibrillation electrocardiograms (Sec.~\ref{sec:aaextr}), and especially in atrial flutter
episodes, typically lies in the sub-Gaussian source subspace. The
ventricular activity sources are usually impulsive
and thus super-Gaussian. If only a few of these sources are desired,
separating the whole mixture would incur an unnecessary computational
cost and, in the case of sequential extraction, an increased source estimation inaccuracy due to error accumulation through
successive deflation stages. A wiser alternative consists of extracting the desired type of sources exclusively.

RobustICA can easily be modified to deal with these situations by targeting a source with
specific kurtosis sign~$\varepsilon$. After computing the roots of the
step-size polynomial, one simply needs to
replace~(\ref{eq:optstep}) by
\begin{equation}\label{eq:optstep2}
\mu\opt = \mathrm{arg}\:\underset{k}{\mathrm{max}}\;\varepsilon\cK(\bw + \mu_k\bg)
\end{equation}
as best root selection criterion. If no source exists with the
required kurtosis sign, the algorithm may converge to a non-extracting
local extrema, but will tend to produce components with maximal 
 or minimal kurtosis from the remaining signal subspace when $\varepsilon = 1$ or $\varepsilon =
 -1$, respectively. The algorithm can also be
run by combining global line
maximizations~(\ref{eq:optstep2}) and~(\ref{eq:optstep}) for sources
with known and unknown kurtosis sign, respectively, in any desired
order.


\subsection{Deflation} \label{sec:defl}

To extract more than one independent component, the Gram-Schmidt-type deflationary
orthogonalization procedure proposed for FastICA \cite{HYV97},
\cite{HYV99}, \cite{HYV01} (see Sec.~\ref{sec:opt}) can also be used in
conjunction with RobustICA under prewhitening, even if prewhitening is
not mandatory for this method. After step S4, the updated
extracting vector is constrained to lie in the orthogonal subspace of the
 extracting vectors previously found [eqn.~(\ref{eq:deflortho})]. In the
 linear regression approach to deflation \cite{TUG97}, after
convergence of the search algorithm the contribution of
the estimated source $\hat{s}$ to the observations is computed via the minimum
mean square error solution to the linear
regression problem $\mathbf{x} = \mathbf{\hat{h}}\hat{s}$.
The observations are then deflated as $\mathbf{x} \leftarrow (\mathbf{x}
- \mathbf{\hat{h}}\hat{s})$ before re-initializing the algorithm in
the search for the next source.
If prewhitening is not performed and the mixture is not unitary, orthogonalization is no
longer an option and an alternative procedure like regression becomes compulsory.


\subsection{A Quick Look at Convergence} \label{sec:conv}

The theoretical study of RobustICA's convergence characteristics in
the general case is beyond the scope of the
present paper. In the real-valued two-signal
scenario, however, the
algorithm converges to the global optimum in a single iteration, even without
prewhitening. The proof relies on the scale invariance property of
contrast~(\ref{eq:kurt}) and follows straightforward
geometrical arguments. Suppose that the initial (non-zero) extracting vector $\mathbf{w}_0$
has an orientation of $\alpha_1$~rad with respect to one of the axis
vectors spanning $\mathbb{R}^2$. In polar coordinates, the gradient
at $\mathbf{w}_0$ can be expressed as 
$$\mathbf{g}_0 = \nabla\cK(\mathbf{w}_0) =
\frac{\partial\cK(\mathbf{w}_0)}{r\partial\theta} \mathbf{u}_\theta +
\frac{\partial\cK(\mathbf{w}_0)}{\partial r} \mathbf{u}_r$$ 
where
$\mathbf{u}_\theta$ and $\mathbf{u}_r$ denote the unit vectors in the
radial and ortho-radial directions, respectively. The radial component can be
computed as
\begin{equation*}
\frac{\partial \cK(\mathbf{w}_0)}{\partial r} = \underset{\alpha \rightarrow 0}{\lim}\:\frac{\cK(\mathbf{w}_0 +
  \alpha\mathbf{u}_r) - \cK(\mathbf{w}_0)}{\alpha} = 0
\end{equation*}
since $\mathbf{w}_0 \propto \mathbf{u}_r$ and the numerator is null for any $\alpha$ by virtue of the
contrast scale invariance. Vector $\mathbf{g}_0$ is orthogonal to
$\mathbf{w}_0$ and its orientation is thus $\alpha_2 = \alpha_1 \pm
\pi/2$~rad. Now, as $\mu$ varies in $\mathbb{R}$,
the orientation of vector $\mathbf{w}_0+\mu\mathbf{g}_0$ spans a
$\pi$-rad interval, which corresponds to the full solution space up to
admissible sign and scale
ambiguities in
the two-signal case. Hence, the optimal step-size technique described
in Sec.~\ref{sec:osskurt} will find the global optimum of the absolute
kurtosis contrast in a single step. Although this result is not easily
generalized to more than two signals, it gives a glimpse of RobustICA's
speed of convergence measured in terms of iterations. By construction
of the algorithm, the OS procedure guarantees at least monotonic convergence of the kurtosis
contrast to a local extremum for any initial condition
(cf.~\cite{REG02, REG03}). Also by construction, consecutive gradient vectors
are orthogonal in the sense that $\re\{\mathbf{g}\tc\mathbf{g}^+\} =
0$, with $\mathbf{g}^+ = \nabla\cK(\mathbf{w}^+)$. This gradient
orthogonality may slow down convergence in high-dimensional extracting
vector spaces.


\subsection{Computational Complexity} \label{sec:comp}

In the literature, complexity is commonly measured in terms of
iterations. Such a measure is unfair in that an algorithm
requiring few iterations to converge may involve heavy computations at
each iteration. The average time taken by an algorithm to achieve a
solution, another complexity measure used in some works \cite{KOL06,
  LI08}, does not take into account the fact that computation time
depends on the actual algorithmic implementation. For instance, when
using the popular Matlab technical computing environment, the
execution time can be considerable reduced if loops are replaced by
vector-wise operations. These observations point out that the number of
real-valued floating point operations (flops) required for an algorithm to reach a solution arises as a
more objective measure of complexity. A flop is considered as a
 product followed by an addition and, in practical
 implementations, would naturally correspond to a
 multiply-and-accumulate cycle in a digital signal processor.
In the signal extraction problem, the
total cost of the extraction can be computed as the product of
the number of iterations, the cost per iteration per source and the
number of extracted sources. The prewhitening stage, if performed, adds
around $2K^2T$ flops ($8K^2T$ in the complex case) to
the total cost when computing the economy singular value decomposition
(SVD) of the data matrix \cite{GOL96}. The complexity
per source per sample is given by the total cost divided by
$KT$.

Table~\ref{tab:comp} summarizes the main
 computations per iteration required by RobustICA and FastICA, for both the
 real-valued and complex-valued scenarios; flop count details can be
 found in~\cite{robusticareport}. Expectations are replaced by sample averages over
 the observed signal block. The sample
  size $T$ is assumed to be sufficiently large, so that only dominant
  terms (with a cost depending on $T$) are considered.
 For the sake of comparison, the complex
 extension of FastICA developed in \cite{BIN00, RIS02} (only valid for
 second-order circular sources) is considered in the corresponding entry of
 Table~\ref{tab:comp}. The CFPA \cite{DOU07} and
 nc-FastICA \cite{NOV08} algorithms [eqn.~(\ref{eq:ncfastica})] have essentially the same
 cost as FastICA in the complex case; it suffices to add an
 initial burden of $L(2L+1)T$~flops due to the computation of the
 pseudo-covariance matrix. The KM-F algorithm \cite{LI08} [eqn.~(\ref{eq:kmf})] takes as many operations
 per iteration
 as RobustICA's gradient computation save for the term
 $\E\{|y|^4\}$, i.e., $(14L + 5)T$ flops.
RobustICA's
 iterations are generally more expensive than FastICA's and its variants. However, as will be demonstrated in
 the next section, each RobustICA iteration is more effective in the search of good
 extraction solutions, so that the overall complexity is actually 
 lower than FastICA's for the same extraction accuracy. Furthermore,
 in some cases FastICA cannot reach RobustICA's accuracy.


\section{Experimental Analysis} \label{sec:exp}

The following experimental analysis evaluates RobustICA's convergence
characteristics, source extraction quality and computational
complexity in
several simulation conditions involving synthetic data. In the real
case (Secs.~\ref{sec:expsaddle}--\ref{sec:expnoise}),
we use the original FastICA algorithm with cubic non-linearity
[eqn.~(\ref{eq:fastica})] as a benchmark, as it offers
the fastest convergence speed among the previously proposed
kurtosis-based source extraction methods. In the complex case
(Sec.~\ref{sec:expcomplex}), we compare RobustICA to recent
FastICA variants capable of dealing with non-circular
sources. The processing
of real data is reported in Sec.~\ref{sec:aaextr}.


\subsection{Robustness to Saddle Points} \label{sec:expsaddle}

The first experiment tests the
comparative convergence characteristics of RobustICA as well as its
robustness to saddle points degrading the performance of the FastICA algorithm for short sample sizes \cite{TIC06}. Independent realizations of two
uniformly distributed sources are mixed through Givens
rotations of random angle $\theta$. The FastICA and RobustICA algorithms are run on the same
mixed data with a sufficiently small termination test $\eta =
0.5\times 10^{-6}$. As a natural measure of extraction quality, we employ the average signal mean square error
(SMSE), a contrast-independent criterion defined as 
\begin{equation} \label{eq:smse}
\text{SMSE} = \frac{1}{K}\sum_{k'= 1}^K \text{SMSE}_{k',\ell'(k')} 
\end{equation}
where $\text{SMSE}_{k,\ell} = \E\{|s_k -
\alpha_\ell\hat{s}_\ell|^2\}$, with $\alpha_\ell =
\E\{s_k\hat{s}_\ell^*\}/\E\{|\hat{s}_\ell|^2\}$. 
Signal pairs $(s_{k'},
\hat{s}_{\ell'(k')})$ are chosen in increasing SMSE
order as $\bigl(k', \ell'(k')\bigr) = \arg\underset{k,
  \ell}{\min}\;\text{SMSE}_{k, \ell}$ and, once selected, are no longer taken into account in the pairing of
the remaining sources. When the source estimation is good enough, this `greedy' algorithm allows an optimal
permutation and scaling of the estimated sources $\{\hat{s}_k\}_{k=1}^K$ before
evaluating the performance index.
In the current setting, the global matrix $\mathbf{G} = \mathbf{W}\t\mathbf{H}$ is also a
 Givens rotation of parameter $\Delta\theta = (\theta -
 \hat{\theta})$, where $\hat{\theta}$ is the rotation angle implicitly
 estimated by the separation methods. 

\begin{figure}[h!]
\centering
\includegraphics[width =
8.5cm]{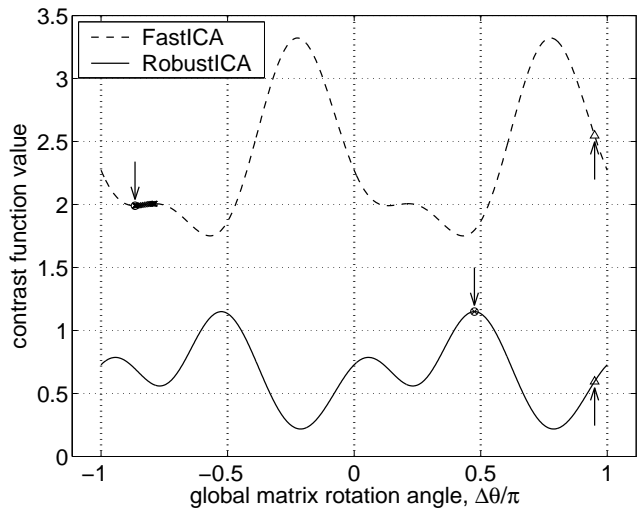}

\smallskip

(a)

\bigskip

\includegraphics[width =
8.5cm]{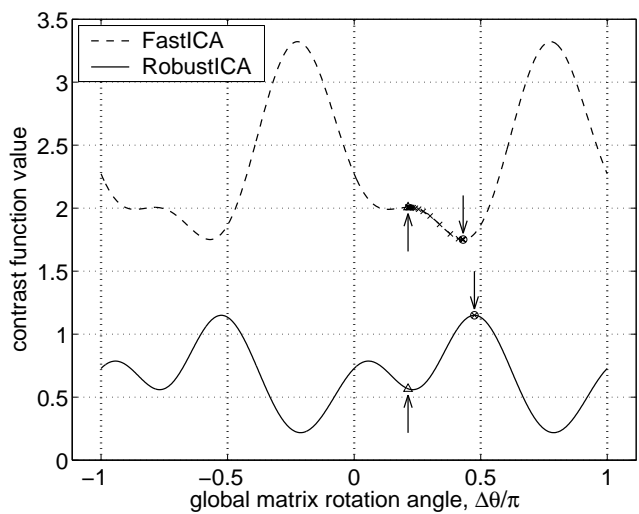}

\smallskip

(b)

\caption{Contrast function values and trajectories for an orthogonal mixture realization of two uniformly distributed sources
  composed of $T = 50$~samples. Dashed line: FastICA's contrast
  function~(\ref{eq:mom}). Solid line: RobustICA's contrast
  function~(\ref{eq:kurt}).  Triangle markers and upward arrows: initial positions. Cross markers: algorithms' solutions after each
  iteration. Round markers and downward arrows: final solutions. Vertical dotted lines: satisfactory
  separation solutions up to sign and permutation. Subplots (a)--(b)
  correspond to two different extracting vector initializations over the same mixture
  realization.}
\label{fig:convergencecontrast}
\end{figure}

\begin{figure}[h]
\centering
\includegraphics[width = 7cm, height = 7cm]{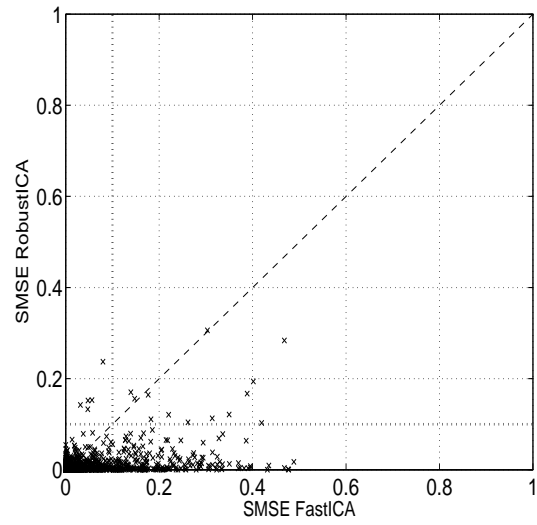}
\caption{Extraction quality scatter plots for the FastICA and RobustICA
  algorithms with random orthogonal mixtures
  of two uniformly distributed sources composed of $T = 50$~samples. Termination parameter $\eta
  = 0.5\times 10^{-6}$, 1000 independent trials.} \label{fig:convergence}
\end{figure}

For a particular signal
 realization, Fig.~\ref{fig:convergencecontrast} plots the contrast
 functions of the respective algorithms [kurtosis~(\ref{eq:kurt}) for
 RobustICA and fourth-order moment~(\ref{eq:mom}) for FastICA] over the optimization
 interval. The small sample size (here 50~samples) smears FastICA's contrast
 function, whose local minima tend to form saddle regions while moving away from the valid separation
 solutions $\Delta\theta = k\pi/2$~rad, $k \in \mathbb{Z}$.
The negative impact of short data length is less manifest for the
kurtosis contrast optimized by RobustICA. 
For the particular initialization shown in Fig.~\ref{fig:convergencecontrast}(a), FastICA gets trapped inside a saddle area between two separation solutions, yielding a final SMSE of
 $-7.8$~dB after 29 iterations. Depending on the initialization, FastICA
 can also converge to the other local minimum with $\mathrm{SMSE} = -13.4$~dB,
 taking up to 24 iterations
 [cf.~Fig.~\ref{fig:convergencecontrast}(b)]. By contrast, RobustICA consistently converges to the
 solutions near $\Delta\theta = \pm\pi/2$~rad with $-22.2$-dB SMSE in a
 single iteration for all initializations, as expected from the theoretical analysis of Sec.~\ref{sec:conv}. 
Figure~\ref{fig:convergence} shows the
 scatter plot of final SMSE values for both methods over 1000
 independent mixture realizations; Table~\ref{tab:convergence} summarizes the average
 performance parameters for different sample size values between~50
 and~150. 

RobustICA provides a faster more
 robust performance, especially for short data sizes. The algorithm's
 robustness to initialization is also demonstrated in
 \cite{eusipco06}. 
These results support
the finite sample analysis of \cite{BER05}, where the kurtosis is
shown to present lower variance than the fourth-order
moment. Similarly, the full expression of the
fourth-order cumulant yields improved extraction performance compared
with the fourth-order moment used in the FastICA algorithm
\cite{BER07}. The optimal step-size technique used in RobustICA
further enhances the finite-sample benefits of the kurtosis contrast. 
%


\subsection{Performance-Complexity Trade-off} \label{sec:qualitycost}

A wireless telecommunications scenario is simulated by considering
noiseless orthogonal random mixtures of $K$ unit-power independent BPSK sources
observed at the output of an $L = K$ element array in signal blocks of
$T$ samples. 
The search for each
extracting vector is initialized with the corresponding canonical basis vector, and
is stopped at a fixed number of iterations. The SMSE performance index~(\ref{eq:smse}) is averaged over 1000
independent random realizations of the sources and the mixing matrix. Extraction solutions are computed
directly from the observed unitary mixtures (`FastICA' and
`RobustICA' legend labels) and after a prewhitening stage
based on the SVD of the observed data matrix (`pw+FastICA',
`pw+RobustICA').

\begin{figure}[h]
\centering
\includegraphics[width = 8.5cm]{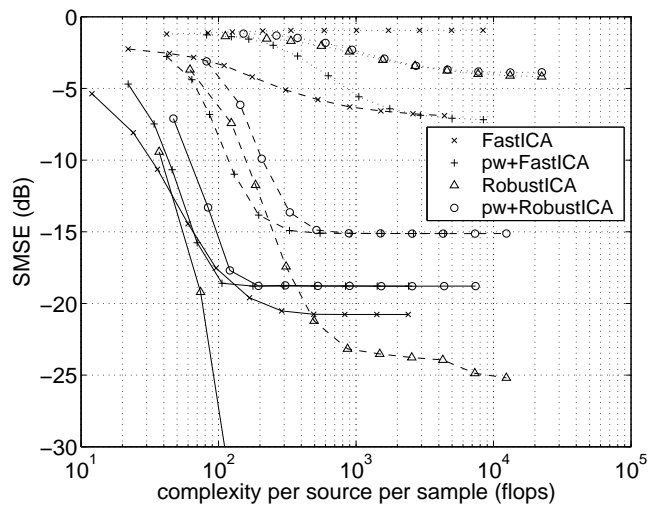}

\caption{Average extraction quality as a function of computational cost for
  different mixture sizes $K$, with signal blocks composed of $T = 150$
  samples and 1000 mixture realizations. Solid lines: $K = 5$. Dashed lines: $K = 10$. Dotted lines:
  $K = 20$.}
\label{fig:qualitycostnsrc}

\end{figure}

\begin{figure}[h]
\centering
\includegraphics[width = 8.5cm]{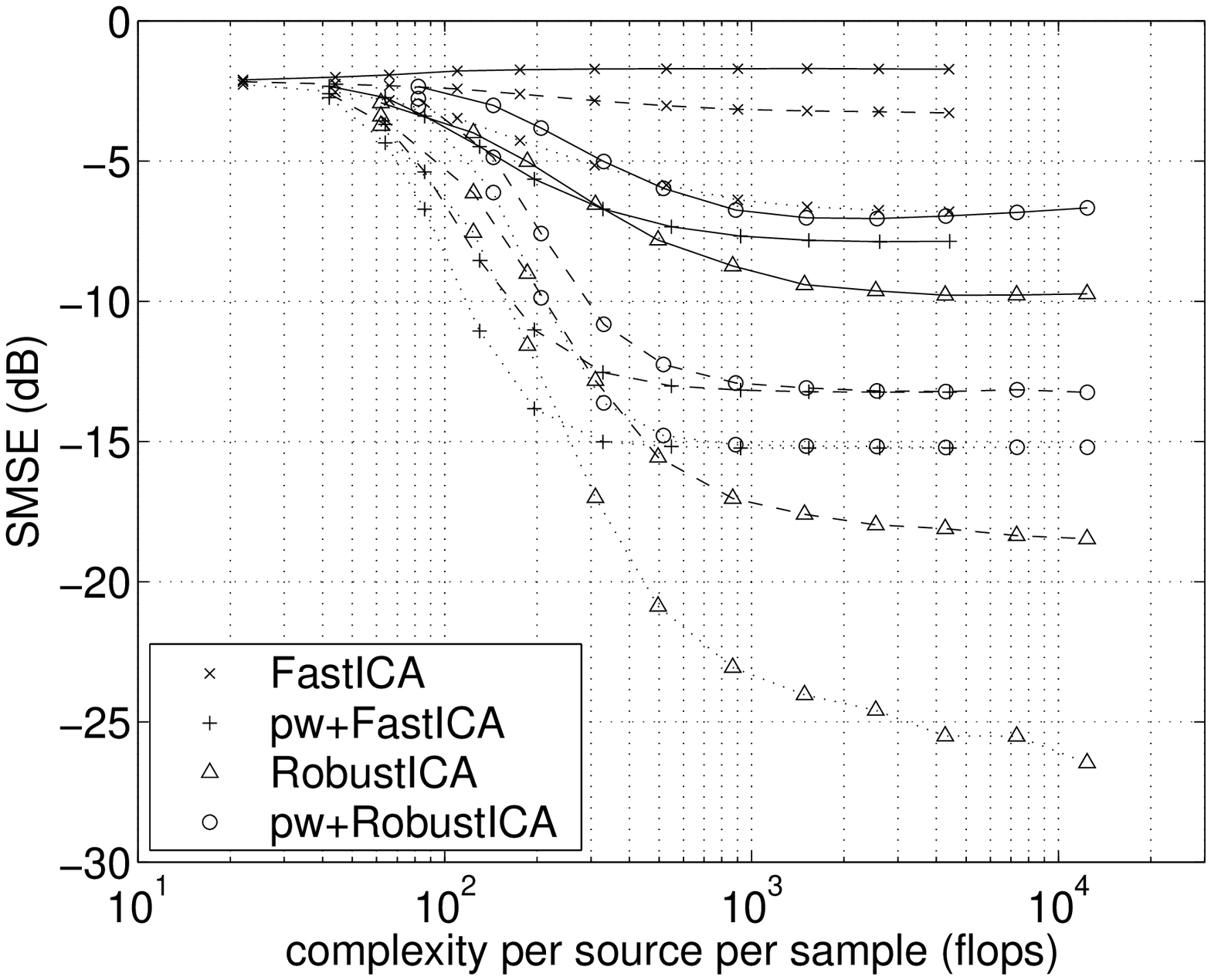}

\caption{Average extraction quality as a function of computational cost for
  different sample sizes $T$, with mixture size $K =
  10$~sources and 1000 mixture realizations. Solid lines: $T = 50$. Dashed lines: $T = 100$. Dotted
  lines: $T = 150$.}
\label{fig:qualitycostT}

\end{figure}

Fig.~\ref{fig:qualitycostnsrc} summarizes the performance-complexity variation obtained for $T = 150$ samples and different values
of the mixture size $K$. The best fastest performance is provided by
RobustICA without prewhitening: a given performance level is achieved
with lower cost or, alternatively, an improved extraction quality is
reached with a given complexity. Although not shown in the plot, the
method gets below the $-60$-dB SMSE level for $K
= 5$~sources in this experiment. The use of prewhitening worsens RobustICA's performance-complexity
trade-off and, due to the finite sample size, imposes the same SMSE
bound for the two
methods. Using prewhitening, FastICA improves considerably and becomes
slightly faster than RobustICA with prewhitening, especially when the mixture size increases.
Fig.~\ref{fig:qualitycostT} displays the quality-cost trade-off for $K =
10$~sources and different block length values. Improved performance
bounds can be achieved by RobustICA if avoiding prewhitening, even for
short data sizes.


\subsection{Efficiency} \label{sec:efficiency}

\begin{figure}[h]
\centering
\includegraphics[width = 8.5cm]{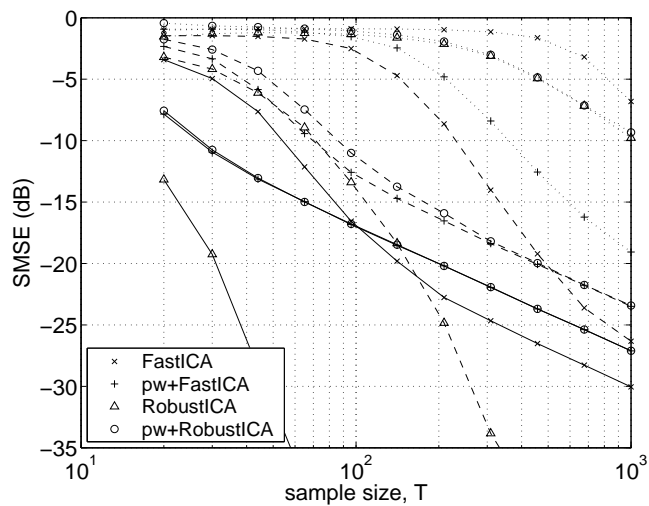}

\caption{Average extraction quality as a function of block length for
  different mixture sizes $K$ with complexity fixed at
  400~flops/source/sample and 1000 mixture realizations. Solid lines: $K = 5$. Dashed lines: $K = 10$. Dotted lines:
  $K = 20$.}
\label{fig:efficiencynsrc}

\end{figure}

We now evaluate the methods' performance for a varying block sample size $T$. Extractions are obtained
by limiting the number of iterations per source, as
explained above. To make the comparison meaningful,
the overall complexity is fixed at 400 flops/source/sample for all tested
methods. Accordingly, since RobustICA is more costly per iteration
than FastICA, it performs fewer iterations per
source. Fig.~\ref{fig:efficiencynsrc} displays the average SMSE curves
for different number of sources $K$. For moderate $K$, RobustICA is
considerably more efficient than the other methods, as shown by the
steeper slope of its curve, achieving the same
extraction performance with much smaller signal blocks. Prewhitening
smoothens FastICA's and RobustICA's performance trends, which become
comparable. As $K$
increases, FastICA with prewhitening becomes more efficient.


\subsection{Performance in the Presence of Noise} \label{sec:expnoise}

\begin{figure}[h]
\centering
\includegraphics[width = 8.5cm]{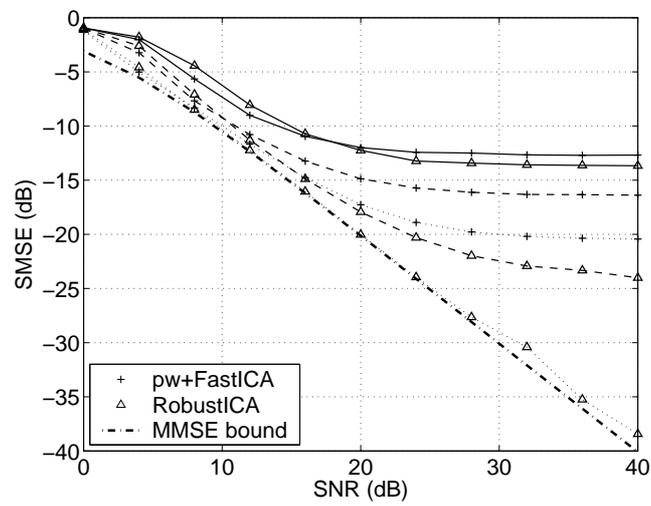}

\caption{Average extraction quality in isotropic additive white
  Gaussian noise with $K = 10$ sources, $T$ samples per source and a complexity fixed at
  400~flops/source/sample and 1000 mixture realizations. Solid lines: $T = 100$. Dashed lines: $T = 200$. Dotted lines: $T = 500$.}
\label{fig:qualitysnr}

\end{figure}

Figure~\ref{fig:qualitysnr} assesses the comparative performance of RobustICA in the presence of noise for $K = 10$~sources, different sample
sizes and a fixed complexity of
400 flops/source/sample. Isotropic additive white Gaussian noise is added to the
observations, with a signal-to-noise ratio (SNR) given by
\begin{equation*}
\mathrm{SNR} =
\frac{\mathrm{trace}(\bH\bH^\mathrm{H})}{\sigma_n^2 L} =
  \frac{1}{\sigma_n^2}
\end{equation*}
where $\sigma_n^2$ denotes the noise power at each sensor output.
The minimum mean square error (MMSE) receiver is shown as a
performance bound for linear detection. RobustICA appears more robust
to additive noise, as it obtains an improved SMSE performance for the same noise
level or, alternatively, it tolerates more noise without sacrificing
performance. At high SNR, RobustICA achieves a lower performance flooring than
FastICA and,
for sufficient sample size, it attains the MMSE bound, employing three times fewer
iterations than the other method in this experiment. Analogous results 
involving noise data are reported in~\cite{ica07}.


\subsection{Complex-Valued Mixtures} \label{sec:expcomplex}

\begin{figure}[h]
\centering
\includegraphics[width =
8.5cm]{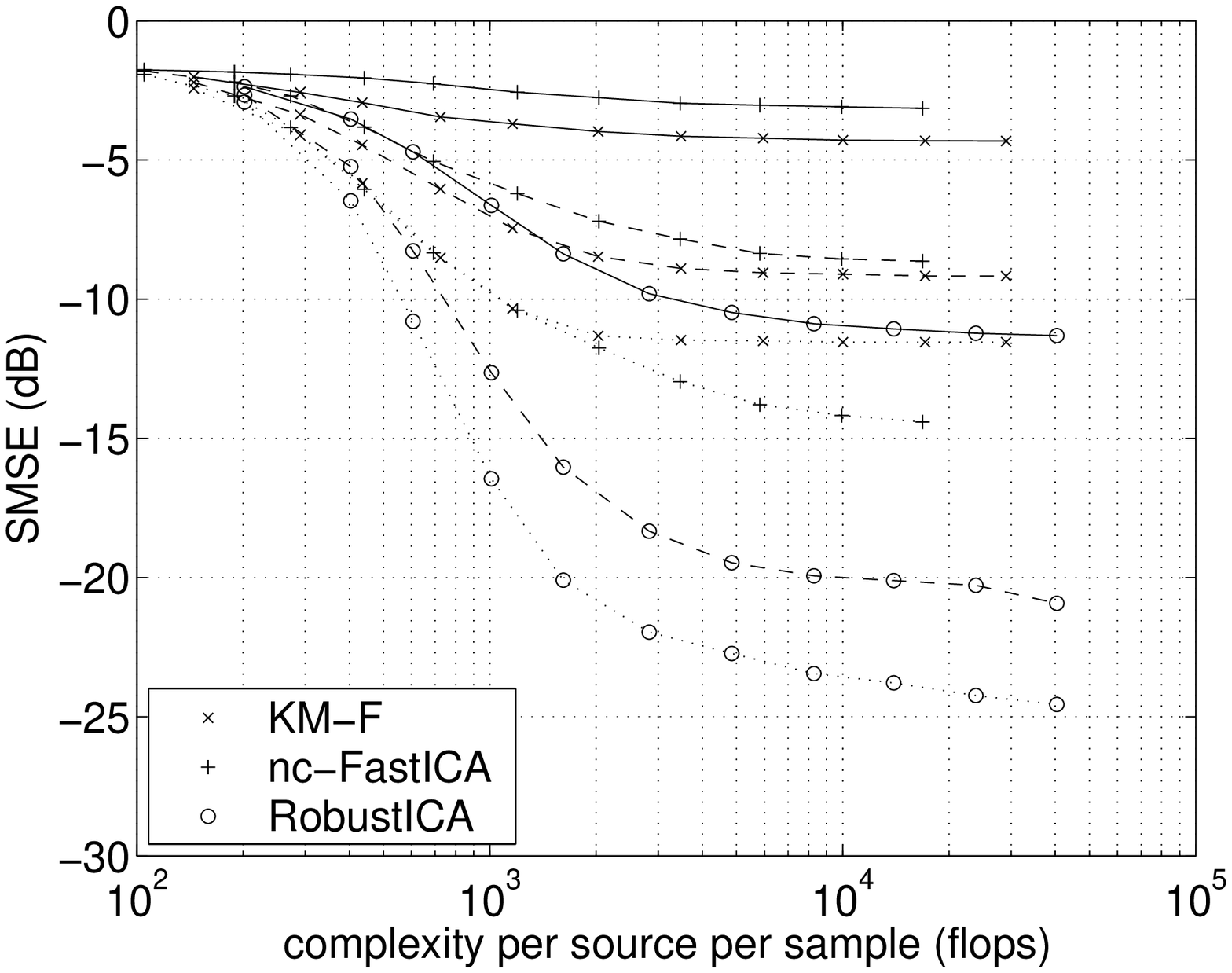}

\smallskip

(a)

\bigskip

\includegraphics[width =
8.5cm]{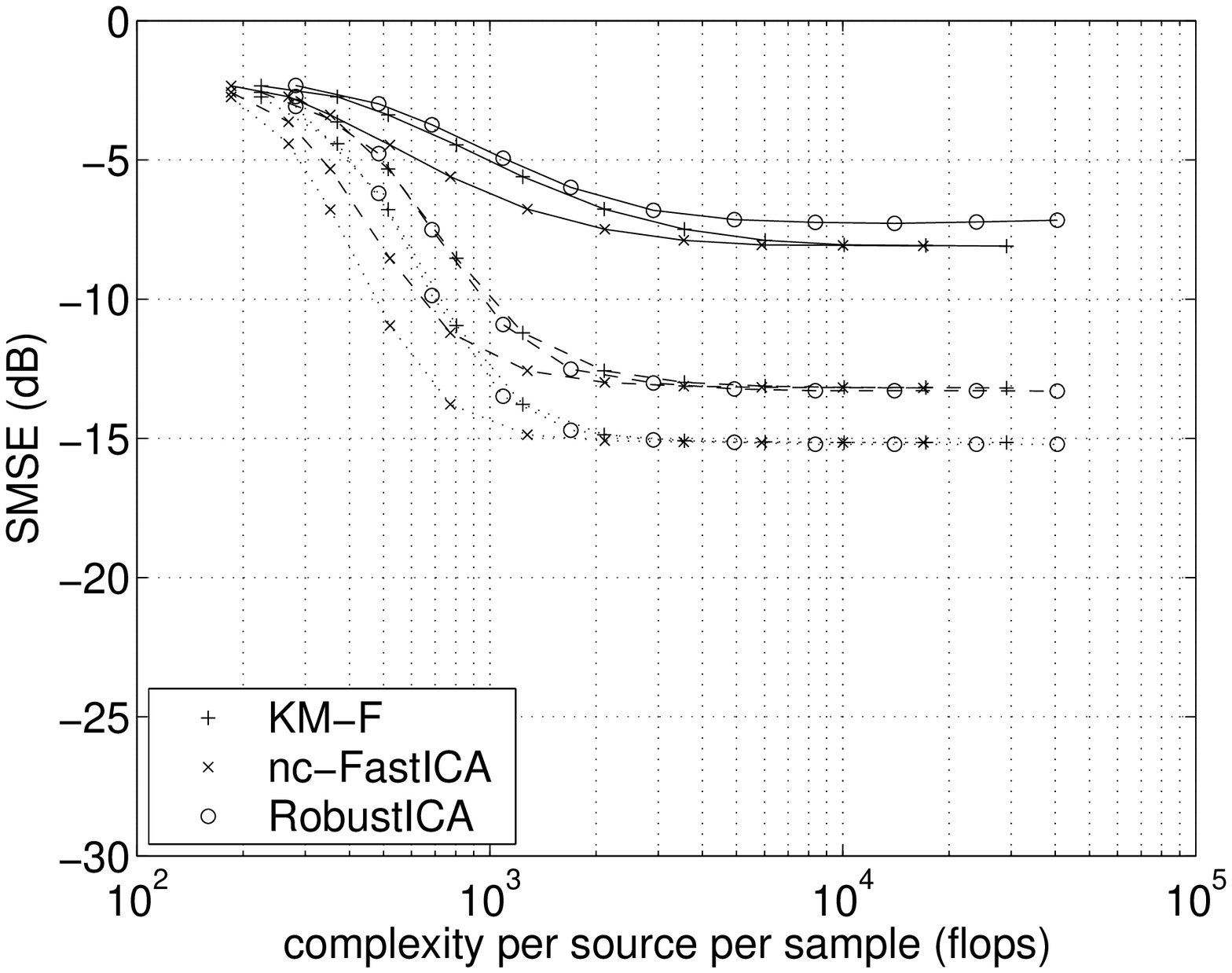}

\smallskip

(b)

\caption{Average extraction quality as a function of computational cost for
  different sample sizes $T$, with mixture size $K =
  10$~sources and 1000 mixture realizations. Solid lines: $T = 50$. Dashed lines: $T = 100$. Dotted
  lines: $T = 150$. (a) Without prewhitening. (b) With prewhitening.}
\label{fig:complexqualitycostT}

\end{figure}

To briefly test RobustICA's performance on complex-valued synthetic mixtures
of non-circular sources,
we repeat the experiment of Sec.~\ref{sec:qualitycost} but using random
unitary mixing matrices. The method is compared with the KM-F
algorithm of \cite{LI08} and the nc-FastICA algorithm of \cite{NOV08}
with kurtosis-based non-linearity, similar to the CFPA algorithm of
\cite{DOU07} (Sec.~\ref{sec:complex}).
The quality-cost trade-off of the three algorithms for different block
sizes is shown in
Fig.~\ref{fig:complexqualitycostT}. Once more, without the performance limitations
imposed by prewhitening, RobustICA proves superior to the
other methods. Performances become similar under prewhitening imposed
to both methods, as FastICA improves whereas RobustICA degrades.


\section{Processing Real Data with RobustICA} \label{sec:aaextr}

Although good performance is obtained with sub-Gaussian
sources \cite{TIC06} as in the above numerical experiments, the use of kurtosis as a general contrast function has been discouraged on
the basis of poor asymptotic efficiency for super-Gaussian sources and
lack of robustness to outliers
\cite{HYV97nnsp}, because the analysis was restricted to FastICA only. This
section reports a biomedical application 
involving non-circular complex strongly super-Gaussian sources where
the kurtosis contrast, optimized by the RobustICA technique, 
shows satisfactory results.

\subsection{Atrial Activity Extraction in Atrial Fibrillation Episodes}

Atrial fibrillation (AF) is the most common cardiac arrhythmia
encountered in clinical practice, affecting up to 10\% of the
population over 70~years of age \cite{FUS06}. The trouble is
characterized by an abnormal atrial electrical activation, whereby the
organized wavefront propagation in normal sinus
rhythm is replaced by several wavelets wandering around the atria in a
disorganized manner. This disorderly electrical activation causes an inefficient atrial mechanical
function and leads to an increased risk of blood-clot formation and stroke. Despite
its incidence, prevalence and risks of serious complications, the
understanding of the generation and self-perpetuation mechanisms of this disease is still
unsatisfactory.

\begin{figure}[h]
\centering
\includegraphics[width = 8.5cm]{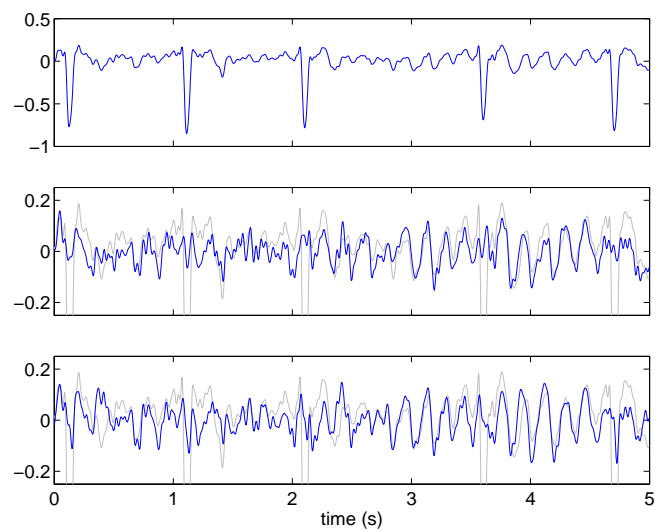}

\caption{Atrial activity extraction in atrial fibrillation ECGs. Top: a 5-second
  segment of lead V1 from the first patient of the database. Middle: AA contribution
  to lead V1 estimated by
  FastICA-SOBI from the 12-lead ECG. Bottom: AA contribution to lead
  V1 estimated by RobustICA from the 12-lead ECG. Only relative
  amplitudes are relevant on the vertical
  axes.}
\label{fig:ecgtime}
\end{figure}

\begin{figure}[h]
\centering
\includegraphics[width = 8.5cm]{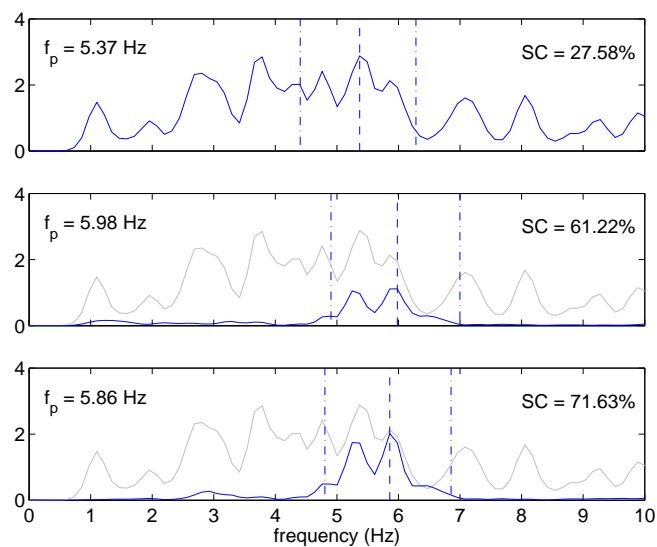}

\caption{Atrial activity extraction in atrial fibrillation
  ECGs. Frequency spectra of the signals shown in Fig.~\ref{fig:ecgtime}. Top: power spectral
  density of signal V1 from the first patient of the database. Middle: power spectral
  density of AA contribution
  to lead V1 estimated by
  FastICA-SOBI from the 12-lead ECG. Bottom: power spectral density of
  AA contribution to lead
  V1 estimated by RobustICA from the 12-lead ECG. Values on the
  left-hand side
  and dashed lines:
  dominant frequency. Values on the right-hand side: spectral
  concentration. Dash-dotted lines: bounds used in the computation
  of spectral concentration. Only relative
  amplitudes are relevant on the vertical
  axes.}
\label{fig:ecgfreq}
\end{figure}

Over recent years, signal processing has helped cardiologists in shedding some light over
AF, as certain features of the atrial activity (AA) signal recorded in
the surface ECG provide information about the arrhythmia. The dominant
frequency of the AA signal is shown to be related to the refractory period of atrial
myocardium cells, and thus to the degree of evolution of the disease
and the probability of spontaneous cardioversion (return to normal
sinus rhythm) \cite{BOL06europace}. The analysis and
characterization of AA from the ECG requires the previous suppression of
interference such as the QRST complex of
ventricular electrical activation (or ventricular activity, VA), artifacts and
noise. Figure~\ref{fig:ecgtime}(top) shows a 5-second segment of precordial
lead V1 from an AF
patient's ECG; its power spectral density, estimated through Welch's
averaged periodogram method as in \cite{tbme05} (averaged 8192-point FFT of
4096-point Hamming-windowed segments with 50\% overlap), is shown in
Fig.~\ref{fig:ecgfreq}(top). The mixture of VA and AA can usually be
perceived in this lead as one of its electrodes lies close to the atria.

A recent approach to AA extraction relies on the observation that AA
and VA can be considered statistically independent phenomena
\cite{tbme04}. Techniques for the
separation of independent signals such as PCA and ICA can
then be applied on the 12-lead ECG to search for the AA source, thus
allowing the reconstruction of AA in all leads free from
VA and other interference. Prior information on the atrial source, in
particular its narrowband character and near-Gaussian behavior, can be exploited to improve AA
extraction performance. In \cite{tbme05}, the kurtosis-based FastICA
method is first applied to extract impulsive interference, essentially
the VA, from the ECG recording. The remaining sources contain mixtures of AA and noise, which, through a
kurtosis-based test, are selected and passed on as inputs to the second-order blind
identification (SOBI) method \cite{BEL97}. Through the joint
approximate diagonalization of the input correlation matrices at
several time lags, SOBI is particularly suited to the separation of
narrowband sources. In this application, the correlation lags are chosen in
accordance with typical AF cycle length values \cite{tbme05}.


\subsection{Application of RobustICA to AA Extraction}

AA is a narrowband signal, so that its frequency-domain
representation is sparse and can thus be considered to stem from an impulsive distribution with high
kurtosis value. Indeed, when mapping
certain signals from the time domain to the frequency or the wavelet domains, the
statistics of the sources tend to become less Gaussian, as observed in
\cite{JAF05tbme} in the context of another biomedical problem. Relying on this simple observation, RobustICA can be applied on
the ECG recording after transformation into the frequency domain. It
is expected that the $f$-domain AA source be found among the first
extracted components (typically those with higher kurtosis values); its time course can then be recovered by
transforming back into the time domain.

This idea is tested on a database of 35 standard ECG segments
recorded from 34~different AF sufferers. Each segment represents an observation
window of around 12~seconds sampled at 1~kHz. Baseline wander and
high-frequency interference are suppressed by zero-phase Chebyshev type-II
highpass and lowpass filters with cut-off frequencies of 0.5~and
30~Hz, respectively. The filtered 12-lead ECG data are then spatially prewhitened before
being passed on to the FastICA-SOBI method of \cite{tbme05}, which
performs all operations in the time domain. Concerning the RobustICA
method, the prewhitened filtered recordings are first transformed into the frequency domain by the
zero-padded 16384-point FFT. The sources extracted in the $f$-domain
are then transformed back to the time domain via the inverse FFT and
truncated to their original length for further
analysis. The AA source is automatically selected as the extracted
component with dominant peak in the interval $[3, 9]$~Hz, the typical
AF frequency band. The percentage of signal power around the
dominant peak, or spectral concentration (SC), has been shown to correlate with AA extraction
quality \cite{tbme05}, and is hence used as a measure of performance. Power
spectra are estimated by Welch's method with the same parameters as in \cite{tbme05}. The same initialization, maximum number of iterations per
source and termination criterion
are used for FastICA and RobustICA.

Figure~\ref{fig:ecgtime}(middle)--(bottom) shows a 5-second segment of the AA
reconstructed by the two methods in lead V1 from the first
patient of the AF ECG database. The corresponding frequency spectra, together with the estimated dominant
peak position and the associated SC values, are shown in
Fig.~\ref{fig:ecgfreq}(middle)--(bottom). As can be seen in the intervals between
successive heartbeats, RobustICA obtains a more accurate estimate
of the AA taking place in lead V1, as quantified by a higher SC
value, requiring a total of 698 iterations or around $2721.8\times 10^6$
flops to separate the whole
mixture (53 iterations or $206.7\times 10^6$ flops if stopped at the AA source, found in the 3rd
extracted component),
for 1178 iterations or $391.1\times 10^6$ flops by FastICA (AA source in the 9th component). Performance
parameters averaged over the whole dataset are
summarized in Table~\ref{tab:ecg}. A cost of about $3.5\times 10^6$
flops due to prewhitening should be added to the complexity figures.
If stopped at the AA source, RobustICA only requires an
average of $62\pm 41$ iterations or $241.3\pm 159.9 \times 10^6$
flops. Remark that, according to
Table~\ref{tab:comp}, RobustICA's cost per
iteration is about an order of magnitude greater than FastICA's in this particular setting.  
These results confirm that RobustICA achieves an improved AA signal
extraction quality with virtually identical dominant frequency
estimate at a comparable complexity relative to the alternative two-stage
technique. As a measure of second-order circularity, the ratio
$|\mathrm{E}\{s^2\}|/\mathrm{E}\{|s|^2\}$ averaged over all $f$-domain sources extracted by
RobustICA is $0.85\pm 0.02$. Since the non-circular second-order
moment $\mathrm{E}\{s^2\}$ cannot be considered to be null, complex-valued extensions of
FastICA such as those proposed in \cite{BIN00, RIS02} would not be
expected to perform well in this context; more recent variants such
as the KM-F and nc-FastICA algorithms \cite{LI08, NOV08}
(Sec.~\ref{sec:complex}) should be more
successful. More importantly, the average kurtosis of the
frequency-domain sources extracted by RobustICA in the frequency
domain is $231$, whereas that of the AA sources equals $731$. These
are strongly super-Gaussian signals.


\section{Conclusions} \label{sec:con}

Kurtosis has long been known to be a valid contrast for independent
source extraction in instantaneous as well as convolutive linear mixtures, whether the sources are real or complex, circular
or non-circular, sub-Gaussian or super-Gaussian, and whether
prewhitening is performed.
The global maximizer of this contrast across the
search direction can be obtained algebraically at each extracting
filter update iteration, giving rise to the
RobustICA method developed in this work. Among
other interesting features naturally inherited from the kurtosis contrast, RobustICA can process real- and complex-valued (possibly non-circular)
sources and does not require prewhitening. As a result, the method is
more tolerant than whitening-based techniques to residual source
correlations likely to appear in short data records. 
In addition,
the optimal step-size approach endows the method with an increased
robustness to initialization and saddle
points, particularly in small observation windows.
The computational complexity required to reach a given source
extraction quality has been put forward as a natural objective measure of
convergence speed for BSS/ICA algorithms. Without
the performance limitations imposed by second-order preprocessing (whitening), RobustICA proves
computationally faster and more efficient than the popular
kurtosis-based FastICA algorithm with asymptotic cubic global
convergence and some of its most recent variants.
RobustICA's ability to process real-world non-circular complex
strongly super-Gaussian signals has been successfully illustrated by the extraction of
atrial activity in atrial fibrillation ECG recordings. In conclusion, the RobustICA method, although conceptually simple,
presents a number of benefits that make it particularly attractive in practical
BSS/ICA settings. Extensions to
convolutive scenarios such as blind SISO and MIMO channel deconvolution are also possible with few modifications. An
illustration of the optimal step-size technique on the kurtosis
contrast in the SISO case is reported in \cite{eusipco05}. The MIMO case
calls essentially for the definition of appropriate deflation procedures along the
lines of \cite{TUG97},
which should be the subject of fresh investigations. More robust cumulant
estimates (see, e.g., \cite[Ch.~5]{NAN99} and references
therein) would increase the
method's ability to handle outliers, and would be another interesting
avenue for the continuation of this work.


\section*{Acknowledgments}
The authors would like to thank the Hemodynamics Department,
Clinical University Hospital, University of Valencia,
Spain, and ITACA-Bioingenier{\'\i}a, Polytechnic University of
Valencia, Spain, for providing the recording database used in
the experiments of Section~V.


\appendix

\section*{Derivation of the Optimal Step-Size Polynomial}

Contrast $\cK$ evaluated at $\bw+\mu\bg$ becomes a function of $\mu$
only, and is given by the rational fraction
\begin{equation} \label{eq:fracratio}
\cK(\mu) = \frac{\E\{|y^+|^4\} - |\E\{(y^+)^2\}|^2}{\E^2\{|y^+|^2\}} - 2 = \frac{P(\mu)}{Q^2(\mu)} - 2
\end{equation}
where $y^+ = y + \mu g$, $y = \bw\tc\bx$, $g = \bg\tc\bx$, $P(\mu) =
P_1(\mu) - |P_2(\mu)|^2$, $P_1(\mu) = \E\{|y^+|^4\}$, $P_2(\mu) = \E\{(y^+)^2\}$
and $Q(\mu) = \E\{|y^+|^2\}$. Let us denote
\begin{equation*}
a = y^2 \qquad b = g^2 \qquad c =
yg \qquad d = \re(yg^*).
\end{equation*}
After some tedious but otherwise
straightforward algebraic manipulations, the above
polynomials turn out to be:
\begin{equation} \label{eq:polpq}
P(\mu) = \sum_{k=0}^4 h_k\mu^k \qquad\qquad Q(\mu) = \sum_{k=0}^2 i_k
\mu^k
\end{equation}
where 
\begin{eqnarray} \notag
h_0 &=& \E\{|a|^2\} - |\E\{a\}|^2, \quad h_1 = 4\E\{|a|d\} - 4\re(\E\{a\}\E\{c^*\})\\ \notag
h_2 &=& 4\E\{d^2\} + 2\E\{|a||b|\} -4|\E\{c\}|^2 - 2\re(\E\{a\}\E\{b^*\})\\  \notag
h_3 &=& 4\E\{|b|d\} - 4\re(\E\{b\}\E\{c^*\}), \quad h_4 = \E\{|b|^2\} - |\E\{b\}|^2\\ 
i_0 &=& \E\{|a|\},\quad i_1 = 2\E\{d\}, \quad i_2 = \E\{|b|\}. 
\end{eqnarray} 
Hence, the derivative of $\cK(\bw+\mu\bg)$ with respect to $\mu$ reads
\begin{equation} \label{eq:dcontrast}
\dot{\cK}(\mu) = \frac{\dot{P}(\mu)Q(\mu) -
  2P(\mu)\dot{Q}(\mu)}{Q^3(\mu)} = \frac{p(\mu)}{Q^3(\mu)}.
\end{equation}
Relating eqns.~(\ref{eq:polpq})--(\ref{eq:dcontrast}), polynomial
$p(\mu)$ is given by eqn.~(\ref{eq:pol}) with
\begin{align*} \notag
a_0 &=  -2h_0 i_1 + h_1 i_0 \quad & a_1 =  -4h_0 i_2 - h_1 i_1 + 2h_2 i_0\\ \notag
a_2 &= -3h_1 i_2 + 3h_3 i_0  \quad & a_3 = -2 h_2 i_2 + h_3 i_1 + 4h_4 i_0 \\ \notag
a_4 &=  -h_3 i_2 + 2h_4 i_1.
\end{align*}
The real parts of the roots of this polynomial are the step-size candidates to be found in
step~S2 of the algorithm (Sec.~\ref{sec:osskurt}). These candidates are then plugged back into
eqns.~(\ref{eq:fracratio})--(\ref{eq:polpq}) to check
which one provides the optimum value of $|\cK(\bw+\mu\bg)|$, or
of $\varepsilon\cK(\bw + \mu\bg)$ if the alternative procedure of
Sec.~\ref{sec:kurtsign} is employed; this is the
optimal step-size sought in step~S3 of the algorithm.



\clearpage

{\Large\bf Tables}

\begin{table}[h!]
\centering
\caption{Computational complexity per iteration in terms of number of real-valued flops
  per iteration for the kurtosis-based FastICA and RobustICA methods. Signal blocks are composed
  of $T$ samples observed at the output of $L$ sensors.} \label{tab:comp}
\begin{tabular}{l|c|c}
Method     & Real Case & Complex Case\\ \hline
FastICA    &  $(2L+2)T$   & $(8L + 4)T$ \\
RobustICA  & $(5L+12)T$   & $(18L+22)T$ \\
\end{tabular} 
\end{table}


\begin{table}[h!]
\centering
\caption{Average performance parameters for the experiments on
  real-valued mixtures of
  Sec.~\ref{sec:expsaddle} and Fig.~\ref{fig:convergence}. Symbol
  $[\cdot]$ denotes the closest integer.} \label{tab:convergence}
\begin{tabular}{c|l|c|c|c|c}
  $T$    & method & SMSE (dB) & iterations & flops $\times 10^3$ & cases with \\
         &        &          & {\scriptsize
         ($[\mathrm{mean}]\pm[\mathrm{std}]$)}&  {\scriptsize
         ($\mathrm{mean}\pm\mathrm{std}$)}& $\mathrm{SMSE} >
  -10$~dB\\ \hline\hline
  50     &  FastICA   & $-11.6$   & $14\pm 56$ & $4.1 \pm 16.8$ & 240\\
       &  RobustICA & $-19.0$   & $1 \pm 0$ & $1.1 \pm 0$ & 18\\ \hline
  100     &  FastICA   & $-14.7$   & $7\pm 6$ & $4.1 \pm 3.8$ & 79\\
       &  RobustICA & $-23.1$   & $1 \pm 0$ & $2.2 \pm 0$ & 0 \\ \hline
 150     &  FastICA   & $-17.0$   & $6\pm 6$ & $5.3\pm 5.1$ & 20\\
       &  RobustICA & $-25.1$   & $1 \pm 0$ & $3.3 \pm 0$ & 0 \\
\end{tabular}
\end{table}



\begin{table}[h!]
\centering
\caption{AA extraction in AF episodes: spectral concentration (SC), position of dominant spectral
  peak ($f_p$), number of iterations, algorithmic complexity and position of estimated AA
  source averaged over the 35~ECG recordings.} \label{tab:ecg}
\begin{tabular}{c||c|c|c|c|c}
         & SC (\%) &  $f_p$ (Hz) & iterations & flops $\times 10^6$ &  AA source position\\
Method   & {\scriptsize (mean $\pm$ std)}&  {\scriptsize (mean
  $\pm$ std)} &  {\scriptsize ($[\mathrm{mean}]\pm[\mathrm{std}]$)} &  {\scriptsize (mean
  $\pm$ std)} & {\scriptsize ($\mathrm{median}\pm[\mathrm{std}]$)}  \\ \hline\hline
FastICA-SOBI & $48.55 \pm 17.06$ & $5.40 \pm 1.18$ & $1245\pm 934$ &
         $406.2\pm 302.8$ & $9 \pm 2$
\\ \hline
RobustICA    &  $55.67 \pm 16.78$ & $5.41 \pm 1.18$ & $202\pm 99$ & $
         786.9 \pm 387.4$ & $3 \pm 2$
\end{tabular}
\end{table}


\clearpage

{\Large\bf Figures}



\end{document}